\documentclass[11pt]{article}
\usepackage[preprint]{acl}
\usepackage{times}
\usepackage{latexsym}
\usepackage[T1]{fontenc}
\usepackage[utf8]{inputenc}
\usepackage{microtype}
\usepackage{inconsolata}
\usepackage{graphicx}
\usepackage{amsmath,amssymb}
\usepackage{booktabs}
\usepackage{xspace}
\usepackage{adjustbox}
\newcommand{\hy}{\hat y}
\newcommand{\ACR}{SpatialDuet\xspace}
\newcommand{\SSt}{SpatialStack-4B\xspace}
\newcommand{\CS}{Cambrian-S-3B\xspace}
\newcommand{\cmark}{$\checkmark$}\newcommand{\xmark}{$\times$}
\newcommand{\gain}[1]{#1}\newcommand{\loss}[1]{#1}
\title{SpatialDuet: Bayesian task routing between a metric expert and a geometry expert for spatial intelligence}

\author{Duy Tran Thanh}

\begin{document}
\maketitle
\begin{abstract}
Embodied assistants and wearable devices need 3D spatial reasoning, and their limited hardware restricts them to small vision-language models (VLMs). Small spatial VLMs acquire different abilities from their training: a model trained on large spatial question sets estimates object and room sizes well, whereas a model with a 3D geometry encoder is better at relative directions and absolute distances. Instead of training a single model for all spatial tasks, as published methods do, we build one system that uses each frozen model for the tasks it handles best. We propose \ACR, which pairs a metric expert with a geometry expert under a Bayesian task router. The metric expert is a data-driven spatial VLM run with three training-free modules: training-aligned sampling, metric-aware minimum Bayes risk decoding and per-type metric calibration. The geometry expert is a VLM with a 3D geometry encoder. For each question type, the router estimates from labelled development questions how likely each expert is to be better, and pools their answers with that probability as weight, in log space for numeric answers and over the option distributions for multiple-choice answers. Extensive experiments on four spatial benchmarks covering video and single images validate \ACR; it outperforms the best published small VLMs by 3.17\% on VSI-Bench and, excluding a model trained on CV-Bench-style data, by 1.44\% on CV-Bench, without training any model weight.
\end{abstract}

\begin{figure*}[t]
\centering
\includegraphics[width=\textwidth]{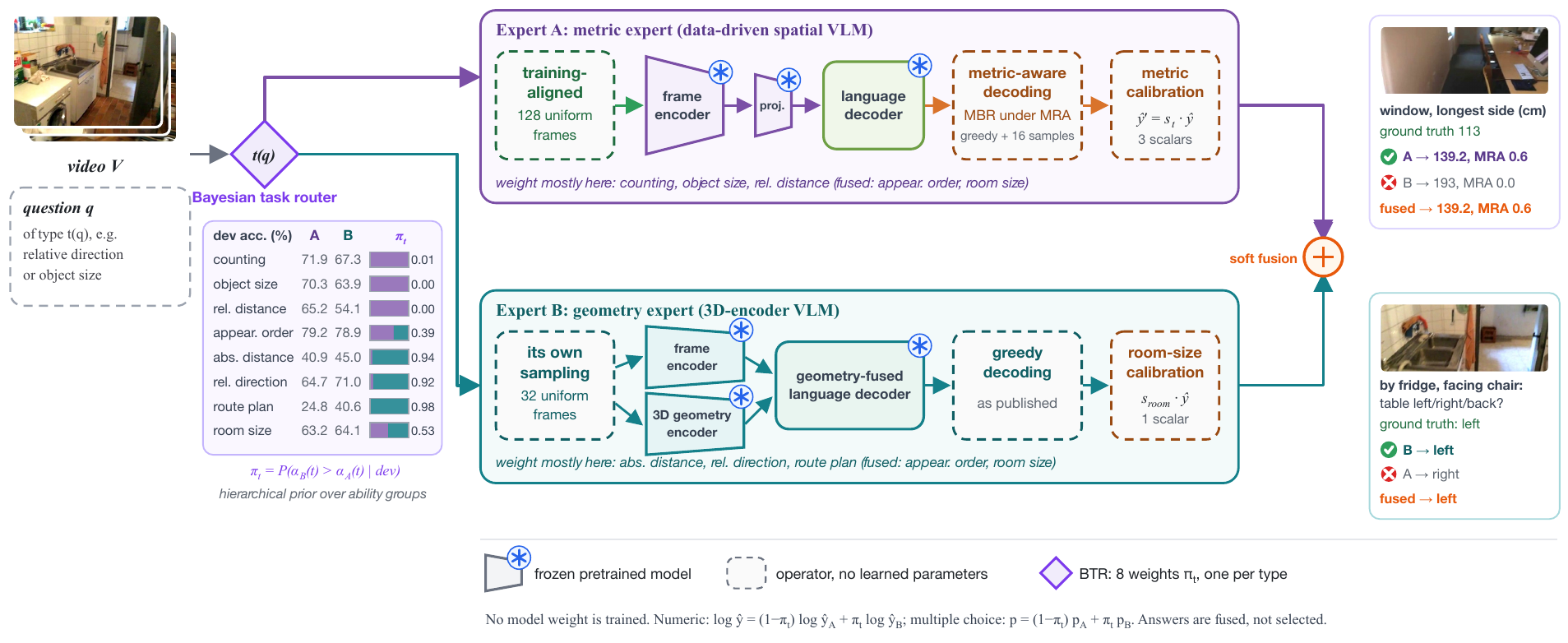}
\caption{\textbf{Overview of \ACR.} A metric expert (A) and a geometry expert (B), both frozen, feed the Bayesian task router, which weights expert B by the probability $\pi_t$ that it is better on question type $t$. Left: development accuracy and $\pi_t$ on VSI-Bench. Right: two development questions with both answers and the router's answer.}
\label{fig:method}
\end{figure*}

\section{Introduction}\label{sec:intro}
Embodied assistants and wearable devices need visual-spatial intelligence~\citep{yang2025thinking}: the ability to tell where objects are, how large they are and how they relate in 3D. Because they run on limited hardware, they rely on small vision-language models (VLMs). Open video VLMs have become strong on spatial benchmarks such as VSI-Bench~\citep{yang2025thinking} along three lines of work. The first fine-tunes on large spatial question sets, as in Cambrian-S~\citep{yang2026cambrians}. The second gives the model 3D geometry, through an added geometry encoder or 3D supervision: VLM-3R~\citep{fan2026vlm3r}, VG-LLM~\citep{zheng2025vgllm}, Spatial-MLLM~\citep{wu2025spatialmllm}, 3DRS~\citep{huang2025threedrs} and SpatialStack~\citep{zhang2026spatialstack}. The third trains spatial reasoning, often with reinforcement learning: ViLaSR~\citep{wu2025vilasr}, SpatialLadder~\citep{li2026spatialladder}, 3DThinker~\citep{chen2026thinkwith3d}, EagleVision~\citep{wan2026eaglevision}, EgoMind~\citep{chen2026egomind} and Object-Centric 3D Rollout~\citep{tang2026oc3d}. Each of these methods trains one model to cover every spatial task.

The first two lines produce models with different abilities. On the VSI-Bench test split, Cambrian-S-3B, trained on a large spatial question set, scores 67.56\% on object size against 61.71\% for SpatialStack-4B, which stacks the features of a VGGT geometry encoder~\citep{wang2025vggt} into its language model. On relative direction the order reverses: 70.54\% for SpatialStack-4B against 62.58\% for Cambrian-S-3B (Table~\ref{tab:pertype}). Choosing, for every question, the better answer of the two experts of Section~\ref{sec:system} would score 74.32\%, against 61.74\% for the better of the two. Two obstacles stand between a single model and this potential. First, the data-driven model is not run the way it was trained and scored: its official evaluation feeds fewer frames than its training, decodes numeric answers greedily, and leaves a per-type scale bias (it overestimates room area by about 20\%). Second, a system must decide, without the ground truth, which model to trust for each kind of question.

We propose \ACR, a training-free system that assigns spatial tasks to two frozen experts according to their abilities (Fig.~\ref{fig:method}). The \emph{metric expert} (expert A) is Cambrian-S-3B run with three modules that align its inference with its training and with the benchmark's metric. Training-aligned sampling (TAS) feeds the 128 uniformly sampled frames of its last training stage. Metric-aware minimum Bayes risk decoding (M-MBR) returns the numeric answer that the model's own samples support most under the benchmark's metric. Per-type metric calibration (PMC) removes the scale bias of each numeric question type. The \emph{geometry expert} (expert B) is SpatialStack-4B, whose geometry encoder supplies 3D cues for relations and directions. The \emph{Bayesian task router} (BTR) decides, for each question type, how much to rely on each expert. It estimates the accuracy gap between the experts on labelled development questions, shrinks noisy estimates toward the gap of the type's ability group, and turns the result into the probability that expert B is better. The two answers are pooled with this probability as weight, in log space for numeric answers and over the option distributions for multiple-choice answers. A type with a clear winner follows that expert, and a type with a small or uncertain gap mixes both, in line with the classic result that a weighted combination of two predictors can be more accurate than the better one~\citep{bates1969combination}.

We evaluate \ACR\ on VSI-Bench, its debiased version, VSTI-Bench~\citep{fan2026vlm3r} and CV-Bench~\citep{tong2024cambrian1}, which cover video and single images. Our contributions are the following.
\begin{itemize}
\item We identify three mismatches between how a data-driven spatial VLM is trained or scored and how it is run, and remove them with three training-free modules (TAS, M-MBR and PMC). Together they raise Cambrian-S-3B from 57.41\% to 61.74\% on the VSI-Bench test split (Section~\ref{sec:metric}).
\item We propose the Bayesian task router, which assigns each spatial question type to the expert whose abilities suit it, using a hierarchical empirical-Bayes estimate of the per-type accuracy gap and format-aware pooling of numeric and multiple-choice answers. Adding the geometry expert through the router raises the score to 64.28\%, 6.87\% above Cambrian-S-3B and 2.54\% above the stronger expert (Section~\ref{sec:router}).
\item \ACR\ outperforms the best published small VLMs by 3.17\% on VSI-Bench and, excluding a model trained on CV-Bench-style data, by 1.44\% on CV-Bench. It also beats the stronger expert on the camera questions of VSTI-Bench, and no model weight is trained (Section~\ref{sec:results}).
\end{itemize}

\section{Related work}\label{sec:related}
\subsection{Spatial video VLMs}
Video multimodal large language models~\citep{openai2024gpt4o,team2024gemini15,chen2024internvl,zhang2024llavavideo,bai2025qwen25vl} remain weak at spatial questions over indoor scenes~\citep{yang2025thinking}. The spatial models reviewed in Section~\ref{sec:intro} narrow this gap by changing the weights of one model: with more spatial training data, with geometry encoders or 3D supervision, or with reasoning trained by reinforcement learning. Each reports a single model per benchmark score, and none assigns tasks across models. Our two experts come from the first two lines: Cambrian-S is trained on a large spatial question set, and SpatialStack stacks VGGT geometry features into the layers of its language model. Recent preprints describe training-free add-ons that give a VLM access to detectors, depth or reconstruction tools~\citep{mou2026visra,izhac2026spacemcp,cho2026spatialclaw,du2026graft,glantz2026seeing}. \ACR\ is training-free as well, and its second source of answers is another spatial VLM.

\subsection{Forecast combination, opinion pools and stacking}
\citet{bates1969combination} showed that two forecasts, combined with weights chosen from their error variances and covariance, can beat both. \citet{granger1984improved} cast the weights as a regression, but estimated weights often do worse than the simple average, a finding that \citet{stock2004combination} documented and that \citet{smith2009simple} explained through the estimation error of the weights. For probability distributions, linear and logarithmic opinion pools have different properties~\citep{genest1986combining}. In machine learning stacked generalization~\citep{wolpert1992stacked} trains a combiner on held-out predictions, and ensembles in general combine outputs by averaging or voting~\citep{dietterich2000ensemble}. The Bayesian task router addresses the estimation problem that these works describe by shrinking per-type estimates toward their ability group.

\subsection{Mixtures of experts, model routing and selective prediction}
Mixtures of experts~\citep{jacobs1991adaptive} learn a gate that sends each input to the experts best suited to it, and sparsely gated layers~\citep{shazeer2017outrageously} made this practical inside large networks. A related line routes whole queries between separately trained language models, either to save cost~\citep{chen2024frugalgpt,ding2024hybrid} or to pick the best model from preference data~\citep{ong2025routellm}. Selective prediction lets a model abstain when it is unsure~\citep{chow1970optimum,geifman2017selective}, and learning to defer sends an input to another decision maker~\citep{madras2018predict,mozannar2020consistent}. The Bayesian task router is a gate over question types: it is estimated in closed form from labelled development questions instead of being trained with the experts. Minimum Bayes risk decoding~\citep{kumar2004minimum} and post-hoc calibration~\citep{platt1999probabilistic,guo2017calibration} are the sources of two modules of the metric expert.

\section{Preliminaries}\label{sec:prelim}
\subsection{Problem formulation}
A spatial question $q$ about an input $V$, a video or a single image, has a type $t(q)$ from a fixed set given by the benchmark, such as object size or relative direction. Its answer is numeric or a choice among options. A numeric answer $\hy>0$ with ground truth $y$ is scored by mean relative accuracy (MRA)~\citep{yang2025thinking},
\begin{equation}
\operatorname{MRA}(\hy,y)=\frac{1}{10}\sum_{\theta\in\Theta}\mathbb 1\!\left[\frac{|\hy-y|}{y}\le1-\theta\right],
\end{equation}
with $\Theta=\{0.50,0.55,\dots,0.95\}$, and a multiple-choice answer by exact match. We use each benchmark's official score, which on VSI-Bench is the macro average over types. We are given two frozen pretrained VLMs and a labelled development split; no model weight is trained.

\subsection{Baseline}\label{sec:baseline}
Our baseline is \CS~\citep{yang2026cambrians}, which combines a SigLIP2 encoder~\citep{tschannen2025siglip2}, a connector and the Qwen2.5-3B language model~\citep{qwen2025qwen25} and is trained on a large spatial question set. With its official evaluation it scores 57.41\% on the VSI-Bench test split. We find four limitations and propose one solution for each.

\textbf{L1: frame mismatch.} The official evaluation takes one frame per second and resamples to 128 frames only when more remain. On the 288 VSI-Bench videos this gives a median of 82 frames (range 22--128), whereas the last training stage of Cambrian-S uses 128 uniformly sampled frames. We propose training-aligned sampling (TAS), which feeds the 128 training frames and raises the score by 1.70\% on the test split (Section~\ref{sec:ablation}).

\textbf{L2: greedy numeric decoding.} The greedy answer to a numeric question is often not the value that the model's own samples agree on. In the example of Fig.~\ref{fig:panels}b the greedy answer is 160\,cm, the value best supported by 16 samples is 145\,cm, and the ground truth is 113\,cm. We propose metric-aware minimum Bayes risk decoding (M-MBR), which returns the value that the samples support most under the benchmark's metric and adds 1.05\%.

\textbf{L3: per-type scale bias.} Numeric answers carry a bias that depends on the question type; room area is overestimated by about 20\% on the development split (Fig.~\ref{fig:panels}c). We propose per-type metric calibration (PMC), which fits one scale per numeric type on the development split and adds 1.57\%.

\textbf{L4: weak relational reasoning.} Without 3D geometry input, \CS\ scores 62.58\% on relative direction and 40.24\% on absolute distance, against 70.54\% and 47.10\% for \SSt, which has a VGGT geometry encoder (Table~\ref{tab:pertype}). We add a geometry expert through the Bayesian task router, which assigns these types to it and adds 2.54\% over the metric expert.

\section{The \ACR\ system}\label{sec:system}
\subsection{Overview}
Fig.~\ref{fig:method} shows the system. Given an input $V$ and a question $q$ of type $t(q)$, each frozen expert $e\in\{A,B\}$ returns an answer $\hy_e$. For multiple-choice questions it also returns a probability $p_e(o)$ over the option letters $o$, taken from the first-step distribution and renormalised over the valid options. The Bayesian task router combines the two answers into one, and no weight of either expert is trained.

The figure shows two real development questions. For the window size, expert A answers 139.2\,cm (MRA 0.6) and expert B answers 193\,cm (MRA 0.0); the router follows expert A because $\pi_{\rm size}=0.000$. For the table direction, expert B answers ``left'' (correct), expert A answers ``right'', and the router answers ``left''. The left panel shows the development accuracy of the two experts per type and the router weight $\pi_t$.

\subsection{Metric expert: \CS\ with three training-free modules}\label{sec:metric}
Expert A keeps every weight of \CS\ unchanged and changes only how the model is run.

\begin{figure*}[t]
\centering
\includegraphics[width=\textwidth]{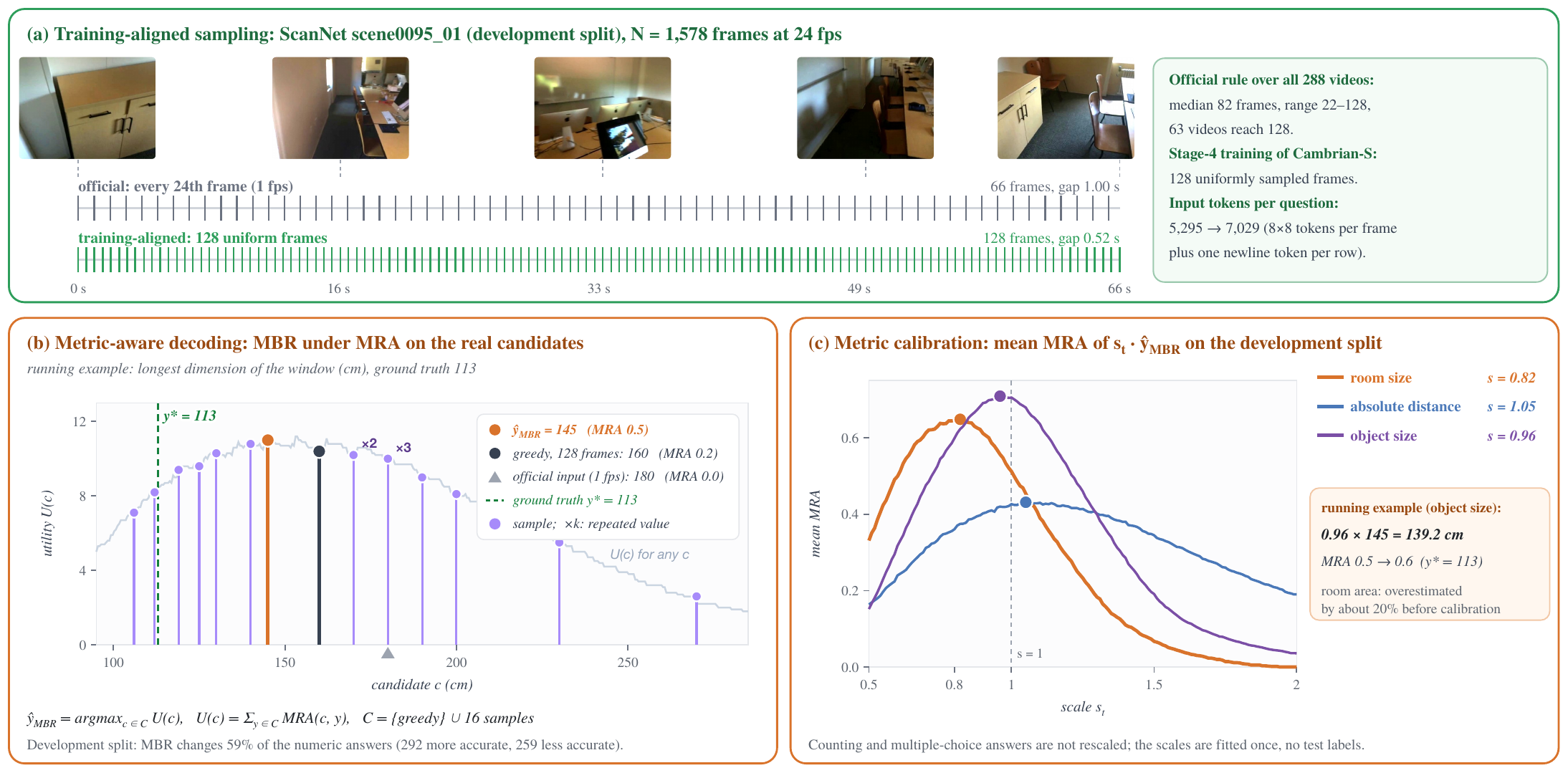}
\caption{\textbf{Real intermediates of the metric expert.} (a) Frame indices under official sampling and TAS. (b) Candidate values and utilities in M-MBR. (c) PMC fit per metric type, with the selected scales marked.}
\label{fig:panels}
\end{figure*}

\textbf{Training-aligned sampling (TAS).} Expert A feeds $\Phi_{\rm align}(V)=\operatorname{linspace}(0,N-1,128)$ for every video of $N$ frames, the sampling of the last training stage~\citep{yang2026cambrians}. For scene0095\_01 (development split, 1,578 frames at 24 frames per second), the official rule takes 66 frames one second apart, and TAS takes 128 frames 0.52\,s apart (Fig.~\ref{fig:panels}a).

\textbf{Metric-aware minimum Bayes risk decoding (M-MBR).} For a numeric question, expert A draws $M=16$ samples (temperature 0.7, nucleus sampling~\citep{holtzman2020curious} with probability 0.95, at most 16 tokens) and forms the candidate set $C=\{\hy_{\rm greedy},y_1,\dots,y_M\}$. It returns the candidate that the others support most under MRA,
\begin{equation}
\begin{gathered}\hy_{\rm MBR}=\arg\max_{c\in C}\ U(c),\\ U(c)=\sum_{y\in C}\operatorname{MRA}(c,\,y),\end{gathered}
\end{equation}
with ties resolved in favour of the greedy answer. This is minimum Bayes risk decoding~\citep{kumar2004minimum}, with the benchmark measure as utility and the model's own samples as pseudo-references. In Fig.~\ref{fig:panels}b, each stem is a distinct candidate value and its height is the utility $U(c)$.

\textbf{Per-type metric calibration (PMC).} For numeric types $t$, expert A rescales the decoded answer, $\hy' = s_t\,\hy_{\rm MBR}$, with
\begin{equation}
s_t=\arg\max_{s\in\mathcal S}\frac{1}{|D_t|}\sum_{i\in D_t}\operatorname{MRA}(s\,\hy_{{\rm MBR},i},\,y_i),
\end{equation}
where $D_t$ is the set of development questions of type $t$ and $\mathcal S$ is the grid from 0.50 to 2.00 in steps of 0.01. Ties go to the value closest to 1, following the spirit of temperature scaling~\citep{guo2017calibration}. Fig.~\ref{fig:panels}c shows the fit for the three metric types of VSI-Bench. Counting and multiple-choice answers are not changed, and a module that does not apply (fixed frames, image input, no numeric type) is skipped; on a benchmark with multiple-choice questions only, expert A is \CS\ itself.

\subsection{Geometry expert: \SSt}\label{sec:geometry}
\SSt~\citep{zhang2026spatialstack} is a 4B spatial model that stacks features of a VGGT geometry encoder~\citep{wang2025vggt} into its language model. We run it frozen with its published protocol (32 uniform frames on VSI-Bench, greedy decoding). On the numeric types of the other benchmarks, and on VSI-Bench room size, its answers are rescaled by per-type scales fitted on its development answers, as in PMC. ``Expert B'' denotes this calibrated version, and ``SS'' denotes \SSt\ exactly as published.

\subsection{Bayesian task router (BTR)}\label{sec:router}
\textbf{Why question types.} A question type is known at test time in these benchmarks, and the experts' strengths are stable within a type (L4 and Fig.~\ref{fig:method}). The router therefore sets one weight $w_t\in[0,1]$ for expert B per type $t$.

\textbf{Format-aware pooling.} Numeric answers $\hy_A,\hy_B>0$ are pooled in log space by a weighted geometric mean, which suits a relative-error metric:
\begin{equation}
\log\hy=(1-w_t)\log\hy_A+w_t\log\hy_B,
\end{equation}
with counts rounded and the other expert's answer used when one has no valid number. Multiple-choice answers are decided by the pooled option distribution, $\hy=\arg\max_o\,(1-w_t)\,p_A(o)+w_t\,p_B(o)$.

\textbf{Hierarchical empirical-Bayes weight.} On labelled development data, for each type $t$ let $\hat d_t$ be the mean of $s_B(q)-s_A(q)$ over its questions, where the scores $s_e(q)$ are exact match, or MRA for numeric questions. Let $\sigma_t$ be the standard deviation of this mean over 200 bootstrap resamples of scenes (or the benchmark's own split unit). The types of a benchmark are grouped into ability groups $g(t)$ taken from the benchmark's own taxonomy; VSI-Bench has three groups (configurational, measurement and spatiotemporal~\citep{yang2025thinking}). Following the empirical-Bayes treatment of \citet{efron1975data}, we assume a normal--normal model, $\hat d_t\sim\mathcal N(d_t,\sigma_t^2)$ and $d_t\sim\mathcal N(\mu_{g(t)},\tau^2)$. Here $\mu_g$ is the precision-weighted mean of $\hat d_t$ over the group's types (for a group with a single type, the mean over all types), and $\tau^2=\max\!\big(0.01^2,\ \operatorname{Var}_t(\hat d_t-\mu_{g(t)})-\overline{\sigma_t^2}\big)$. The posterior of $d_t$ is normal with
\begin{equation}
\begin{gathered}\tilde d_t=\frac{\tau^2\hat d_t+\sigma_t^2\mu_{g(t)}}{\tau^2+\sigma_t^2},\\ v_t=\frac{\tau^2\sigma_t^2}{\tau^2+\sigma_t^2},\\ w_t=\pi_t=\Phi\big(\tilde d_t/\sqrt{v_t}\big),\end{gathered}
\end{equation}
so $\pi_t$ is the probability that expert B is better on type $t$. A large gap measured on many questions gives $\pi_t$ near 0 or 1, and the type follows one expert. A small or noisy gap is shrunk toward the group mean and gives $\pi_t$ near 0.5, and the type mixes both answers. The router has no parameter fitted by optimisation: its weights follow in closed form from the development gaps.

\section{Experiments}\label{sec:results}
\subsection{Setup}
Table~\ref{tab:suite} lists the benchmarks. VSI-Bench~\citep{yang2025thinking} is split by scene into development (2,671 questions) and test (2,459) halves. VSI-Bench-Debiased~\citep{brown2026benchmark} keeps 2,362 questions that a text-only model cannot answer from non-visual patterns; we re-score existing predictions on its test part (1,160 questions). VSTI-Bench~\citep{fan2026vlm3r} has 312 ScanNet~\citep{dai2017scannet} validation videos and 5,736 questions of nine types, three object relations and six camera types, with the labels after the dataset's erratum of 13 July 2026. CV-Bench~\citep{tong2024cambrian1} has 2,638 single-image questions, split by source image into 1,360 development and 1,278 test questions. The development split of each benchmark is used to fit the calibration scales and the router weights, and the test questions of the VSTI-Bench camera types and of CV-Bench were scored once. All runs use two NVIDIA T4 GPUs in fp16 with sdpa attention. Accuracies and differences are in percent, and differences are absolute. Confidence intervals in the tables are paired 95\% bootstrap intervals~\citep{efron1979bootstrap} (1,000 resamples) over scenes, videos or source images.

\begin{table*}[t]
\centering\footnotesize
\caption{\textbf{The benchmark suite.} Types are question types. The metric is each benchmark's official score. VSTI-Bench dev/test: six camera types in questions, object relations in videos. Only the test part of VSI-Bench-Debiased is used.}
\label{tab:suite}
\setlength{\tabcolsep}{3pt}
\adjustbox{max width=\textwidth}{%
\begin{tabular}{@{}llllcllll@{}}
\toprule
Benchmark & Venue & Input & Questions & Types & Answer formats & Split unit & Dev / test\\
\midrule
VSI-Bench & CVPR'25 & video & 5,130 & 8 & numeric, multiple choice & scene & 2,671 / 2,459 q\\
VSI-Bench-Debiased & COLM'26 & video & 2,362 & 8 & numeric, multiple choice & scene & -- / 1,160 q\\
VSTI-Bench & CVPR'26 & video, 32 frames & 5,736 & 9 & numeric, multiple choice & video & 838 / 3,253 q (camera); 46 / 186 videos (object relations)\\
CV-Bench & NeurIPS'24 & single image & 2,638 & 4 & multiple choice & source image & 1,360 / 1,278 q\\
\bottomrule
\end{tabular}}
\end{table*}

\subsection{Comparison with the state of the art}
\textbf{VSI-Bench.} Table~\ref{tab:sota} compares \ACR\ with published models of at most 4B parameters. Published rows are copied from the original papers: GPT-4o and Gemini-1.5 Pro from VSI-Bench~\citep{yang2025thinking}, Spatial-MLLM from Table 5 of its paper~\citep{wu2025spatialmllm}, and the others from their own papers. On all 5,130 questions, with development questions routed by weights fitted on the other development fold, \ACR\ scores 64.07\%. This is 3.17\% above the best published small model, SpatialStack (60.9\%), and 6.90\% above \CS\ run with the same evaluation code. \ACR\ is best on six of the eight types, with the largest gains over \CS\ on room size (+22.05\%), relative direction (+10.80\%) and route planning (+9.28\%). Larger fine-tuned models score higher, up to 67.5\% (Cambrian-S-7B~\citep{yang2026cambrians} and SpatialStack-5B~\citep{zhang2026spatialstack}); \ACR\ uses a 3B and a 4B model.

\begin{table*}[t]
\centering\footnotesize
\caption{\textbf{Comparison with published models of at most 4B parameters on VSI-Bench (\%).} Published rows are copied from the papers, so protocols differ. $\dagger$: our run on all 5,130 questions (descriptive). Best per column in bold, second best underlined.}
\label{tab:sota}
\setlength{\tabcolsep}{2.4pt}
\adjustbox{max width=\textwidth}{%
\begin{tabular}{@{}llcc cccc cccc c@{}}
\toprule
 & & & & \multicolumn{4}{c}{Numerical Question} & \multicolumn{4}{c}{Multiple-Choice Question} & \\
\cmidrule(lr){5-8}\cmidrule(lr){9-12}
Methods & Venue & Size & \begin{tabular}[c]{@{}c@{}}Spat.\\train.\end{tabular} & \begin{tabular}[c]{@{}c@{}}Obj.\\Cnt.\end{tabular} & \begin{tabular}[c]{@{}c@{}}Abs.\\Dist.\end{tabular} & \begin{tabular}[c]{@{}c@{}}Obj.\\Size\end{tabular} & \begin{tabular}[c]{@{}c@{}}Room\\Size\end{tabular} & \begin{tabular}[c]{@{}c@{}}Rel.\\Dist.\end{tabular} & \begin{tabular}[c]{@{}c@{}}Rel.\\Dir.\end{tabular} & \begin{tabular}[c]{@{}c@{}}Route\\Plan\end{tabular} & \begin{tabular}[c]{@{}c@{}}Appr.\\Order\end{tabular} & Avg.\\
\midrule
\multicolumn{13}{l}{\textit{Proprietary models (reference)}}\\
GPT-4o~\cite{openai2024gpt4o} & -- & -- & \xmark & 46.2 & 5.3 & 43.8 & 38.2 & 37.0 & 41.3 & 31.5 & 28.5 & 34.0\\
Gemini-1.5 Pro~\cite{team2024gemini15} & -- & -- & \xmark & 56.2 & 30.9 & 64.1 & 43.6 & 51.3 & 46.3 & 36.0 & 34.6 & 45.4\\
\midrule
\multicolumn{13}{l}{\textit{Published small models ($\le$4B)}}\\
SpatialStack~\cite{zhang2026spatialstack} & CVPR'26 & 4B & \cmark & 69.2 & \underline{45.4} & 63.0 & \underline{63.2} & 57.9 & \underline{68.4} & \underline{40.2} & \underline{79.6} & \underline{60.9}\\
3DThinker~\cite{chen2026thinkwith3d} & CVPR'26 & 3B & \cmark & -- & -- & -- & -- & -- & -- & -- & -- & 59.1\\
Cambrian-S~\cite{yang2026cambrians} & ICLR'26 & 3B & \cmark & \underline{70.7} & 40.6 & \underline{68.0} & 46.3 & \textbf{64.8} & 61.9 & 27.3 & 78.8 & 57.3\\
Obj.-Centric 3D Rollout~\cite{tang2026oc3d} & AAAI'26 & 3B & \cmark & 63.2 & 34.1 & 57.4 & 46.7 & 39.6 & 45.5 & \textbf{44.3} & 49.8 & 47.5\\
VG-LLM~\cite{zheng2025vgllm} & NeurIPS'25 & 4B & \cmark & 66.0 & 37.8 & 55.2 & 59.2 & 44.6 & 45.6 & 33.5 & 36.4 & 47.3\\
Spatial-MLLM~\cite{wu2025spatialmllm} & NeurIPS'25 & 4B & \cmark & 65.3 & 34.8 & 63.1 & 45.1 & 41.3 & 46.9 & 33.5 & 46.3 & 47.0\\
SpatialLadder~\cite{li2026spatialladder} & ICLR'26 & 3B & \cmark & 63.5 & 34.3 & 61.7 & 43.9 & 45.4 & 44.8 & 35.6 & 36.4 & 45.7\\
Qwen2.5-VL$^\dagger$~\cite{bai2025qwen25vl} & -- & 3B & \xmark & 30.9 & 12.0 & 19.7 & 29.9 & 34.8 & 44.8 & 31.4 & 30.3 & 29.21\\
\midrule
\multicolumn{13}{l}{\textit{Same evaluation code, all 5,130 questions (ours)}}\\
\CS$^\dagger$ (baseline) & ICLR'26 & 3B & \cmark & 70.64 & 40.66 & 67.95 & 46.42 & 64.51 & 61.74 & 26.80 & 78.64 & 57.17\\
\SSt$^\dagger$ (published protocol) & CVPR'26 & 4B & \cmark & 67.58 & 46.04 & 62.84 & 61.22 & 55.21 & 70.86 & 35.05 & 78.64 & 59.68\\
\textbf{\ACR\ (ours)} & -- & 3B + 4B & \cmark & \textbf{71.75} & \textbf{47.21} & \textbf{70.29} & \textbf{68.47} & \underline{64.37} & \textbf{72.54} & 36.08 & \textbf{81.88} & \textbf{64.07}\\
\textit{Gain over the baseline} & & & & \gain{+1.11} & \gain{+6.55} & \gain{+2.34} & \gain{+22.05} & \loss{$-$0.14} & \gain{+10.80} & \gain{+9.28} & \gain{+3.24} & \gain{+6.90}\\
\bottomrule
\end{tabular}
}
\end{table*}

\textbf{CV-Bench.} Table~\ref{tab:sota_cv} gives the same comparison on CV-Bench. On all 2,638 questions, with development rows cross-fitted over two folds of source images, \ACR\ scores 82.64\%, which is 1.44\% above the best published small model, SpatialStack (81.2\%). It is best on both the 2D (77.79\%) and the 3D (87.50\%) averages. MM-Spatial-3B~\citep{daxberger2025mmspatial} scores 91.2\%, but it is fine-tuned on data generated with CV-Bench-style templates, so we do not include it in Table~\ref{tab:sota_cv}.

\begin{table*}[t]
\centering\footnotesize
\caption{\textbf{Comparison with published models of at most 4B parameters on CV-Bench (\%).} Published rows are copied from the papers; Avg.\ is the mean of 2D and 3D. Spatial-MLLM as evaluated by 3DThinker. The last block uses our evaluation code on all 2,638 questions (descriptive). Best per column in bold, second best underlined, among the published rows and \ACR.}
\label{tab:sota_cv}
\setlength{\tabcolsep}{3.5pt}
\adjustbox{max width=\textwidth}{%
\begin{tabular}{@{}lllccccccc@{}}
\toprule
 & & & \multicolumn{3}{c}{2D} & \multicolumn{3}{c}{3D} & \\
\cmidrule(lr){4-6}\cmidrule(lr){7-9}
Methods & Venue & Size & Count & Relation & Avg. & Depth & Distance & Avg. & Avg.\\
\midrule
\multicolumn{10}{l}{\textit{Published small models ($\le$4B)}}\\
SpatialStack~\cite{zhang2026spatialstack} & CVPR'26 & 4B & -- & -- & \underline{75.4} & -- & -- & \underline{87.0} & \underline{81.2}\\
3DThinker~\cite{chen2026thinkwith3d} & CVPR'26 & 3B & -- & -- & -- & -- & -- & -- & 78.4\\
Cambrian-S~\cite{yang2026cambrians} & ICLR'26 & 3B & -- & -- & -- & -- & -- & -- & 75.2\\
Spatial-MLLM~\cite{wu2025spatialmllm} & NeurIPS'25 & 4B & -- & -- & -- & -- & -- & -- & 73.8\\
SpatialLadder~\cite{li2026spatialladder} & ICLR'26 & 3B & -- & -- & 72.4 & -- & -- & 74.9 & 73.7\\
\midrule
\multicolumn{10}{l}{\textit{Same evaluation code, all 2,638 questions (ours)}}\\
\CS\ (baseline) & ICLR'26 & 3B & 70.56 & 82.77 & 75.48 & 77.17 & 75.33 & 76.25 & 75.86\\
\SSt\ (published protocol) & CVPR'26 & 4B & 59.14 & 86.77 & 70.78 & 88.17 & 86.67 & 87.42 & 79.10\\
\textbf{\ACR\ (ours)} & -- & 3B + 4B & 70.69 & 87.54 & \textbf{77.79} & 88.00 & 87.00 & \textbf{87.50} & \textbf{82.64}\\
\textit{Gain over the baseline} & & & \gain{+0.13} & \gain{+4.77} & \gain{+2.31} & \gain{+10.83} & \gain{+11.67} & \gain{+11.25} & \gain{+6.78}\\
\bottomrule
\end{tabular}
}
\end{table*}

\subsection{Ablation study}\label{sec:ablation}
Table~\ref{tab:components} adds one module at a time on VSI-Bench. On the test split, TAS raises the score of \CS\ from 57.41\% to 59.11\% (+1.70\%). PMC raises it to 60.68\% (+1.57\%), and M-MBR to 61.74\% (+1.05\%). Together the three modules make expert A, which is 4.33\% above \CS. Adding the geometry expert through the Bayesian task router raises the score to 64.28\%, 2.54\% above expert A and 6.87\% above \CS. Every module raises the score on both the development and the test split.

\begin{table}[t]
\centering\footnotesize
\caption{\textbf{Component ablation on VSI-Bench (\%).} Each row adds one part to the row above. Difference is against \CS\ official. The PMC row uses scales fitted on greedy answers, and the M-MBR row refits them on the decoded answers. Development rows are cross-fitted where a router is used.}
\label{tab:components}
\setlength{\tabcolsep}{3pt}
\begin{adjustbox}{max width=\columnwidth}
\begin{tabular}{@{}lcccc@{}}
\toprule
 & \multicolumn{2}{c}{Test (2,459)} & \multicolumn{2}{c}{Dev (2,671)}\\
\cmidrule(lr){2-3}\cmidrule(lr){4-5}
Configuration & Acc. & Difference & Acc. & Difference\\
\midrule
\CS, official (baseline) & 57.41 & -- & 56.87 & --\\
\quad + TAS & 59.11 & +1.70 & 58.45 & +1.58\\
\quad + PMC & 60.68 & +3.27 & 59.48 & +2.61\\
\quad + M-MBR (= expert A) & 61.74 & +4.33 & 60.03 & +3.16\\
\quad + geometry expert via BTR (= \ACR) & \textbf{64.28} & +6.87 & \textbf{63.89} & +7.02\\
\bottomrule
\end{tabular}
\end{adjustbox}
\end{table}

Table~\ref{tab:pertype} shows where the gains come from. Expert A improves most on room size (from 47.78\% to 67.50\%), where L3 is largest. \SSt\ is better than expert A on absolute distance, relative direction and route planning, the types where L4 applies. \ACR\ is the best or tied best on seven of the eight types and exceeds both experts on appearance order, absolute distance, relative direction, room size and route planning. It is 5.46\% above \SSt.

\begin{table}[t]
\centering\footnotesize
\caption{\textbf{VSI-Bench test split, accuracy per type (\%).} CS: \CS\ official; A: metric expert; SS: \SSt\ as published; BTR: \ACR\ with the Bayesian task router. Difference is BTR minus the column, with paired 95\% intervals over scenes. Best per row in bold.}
\label{tab:pertype}
\setlength{\tabcolsep}{4pt}
\begin{adjustbox}{max width=\columnwidth}
\begin{tabular}{@{}lcccc@{}}
\toprule
Type & CS & A & SS & BTR\\
\midrule
Appearance order & 81.56 & 84.04 & 78.37 & \textbf{87.94}\\
Absolute distance & 40.24 & 45.44 & 47.10 & \textbf{49.12}\\
Object counting & 71.28 & \textbf{71.68} & 67.88 & \textbf{71.68}\\
Relative direction & 62.58 & 64.84 & 70.54 & \textbf{72.26}\\
Relative distance & 62.50 & \textbf{64.19} & 56.76 & 63.85\\
Object size & 67.56 & \textbf{70.40} & 61.71 & \textbf{70.40}\\
Room size & 47.78 & 67.50 & 59.17 & \textbf{68.89}\\
Route planning & 25.81 & 25.81 & 29.03 & \textbf{30.11}\\
\midrule
Macro average & 57.41 & 61.74 & 58.82 & \textbf{64.28}\\
Difference of BTR & +6.87 & +2.54 & +5.46 & \\
95\% CI & [4.88, 9.15] & [0.66, 4.68] & [3.28, 7.84] & \\
\bottomrule
\end{tabular}
\end{adjustbox}
\end{table}

\subsection{How does the router assign tasks?}
Fig.~\ref{fig:method} (left) lists the router weights fitted on the VSI-Bench development split. The router gives expert B almost no weight on counting ($\pi_t=0.01$), object size (0.00) and relative distance (0.00), the measurement types where expert A is stronger. It gives expert B most of the weight on absolute distance (0.94), relative direction (0.92) and route planning (0.98), where geometry matters. On appearance order (0.39) and room size (0.53) the development gap is small, and the router mixes the two experts. These two types gain most from pooling: on the test split \ACR\ scores 87.94\% on appearance order, 3.90\% above the better expert, and 68.89\% on room size, 1.39\% above the better expert (Table~\ref{tab:pertype}). On CV-Bench the router follows expert A on counting ($\pi_t=0.000$) and expert B on relation, depth and distance ($\pi_t\ge0.943$), which again matches the abilities of the two models.

\subsection{Qualitative results}
\begin{figure*}[t]
\centering
\includegraphics[width=\textwidth]{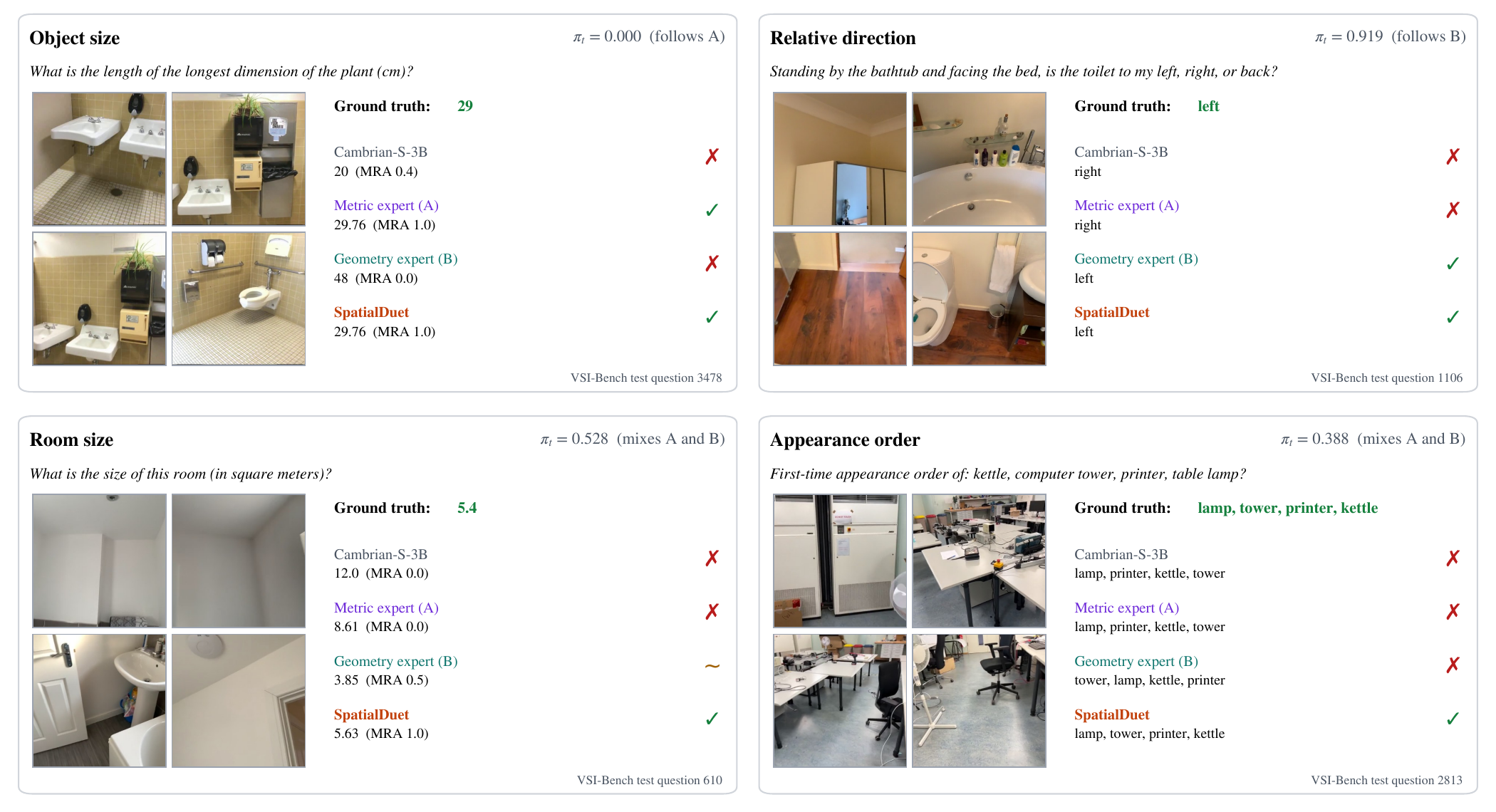}
\caption{\textbf{Qualitative results on the VSI-Bench test split.} Each panel shows four frames of the input video, the answers of \CS, the metric expert (A), the geometry expert (B) and \ACR, and the router weight $\pi_t$ of the question type. \checkmark: correct (MRA $\ge$ 0.8 for numeric answers); $\times$: wrong (MRA $<$ 0.5); $\sim$: partly correct. ``Lamp'' and ``tower'' abbreviate table lamp and computer tower.}
\label{fig:qual}
\end{figure*}

Fig.~\ref{fig:qual} shows four questions from the VSI-Bench test split. On object size the router follows the metric expert ($\pi_t=0.000$), whose answer of 29.76\,cm for a 29\,cm plant is close to exact, while the geometry expert answers 48\,cm. On relative direction it follows the geometry expert ($\pi_t=0.919$), the only one that answers ``left''. On room size, pooling the answers of the two experts (8.61 and 3.85\,m$^2$) in log space gives 5.63\,m$^2$ for a 5.4\,m$^2$ room, closer than either expert. On appearance order neither expert ranks the correct option first, but the pooled option distribution does.

\subsection{Generalization to other benchmarks}
\label{sec:across}
Table~\ref{tab:summary} summarises \ACR\ against \CS\ and the stronger expert on the three test sets.

\textbf{VSTI-Bench.} The six camera types have 3,253 test questions on 250 videos; the metric is the macro average over types, and the stronger expert is A. \ACR\ scores 57.23\%, which is 1.36\% above expert A (55.87\%) and 3.42\% above \CS. On the official VSTI-Bench overall score (mean of all nine types, test videos), it scores 62.73\%, against 57.98\% for \CS\ (+4.75\%) and 61.16\% for \SSt\ (+1.56\%). Table~\ref{tab:vsti} gives the scores of all nine types. The router follows expert B on camera displacement, camera--object absolute distance and the object relations, and expert A on the camera--object relative distances. \ACR\ is best on six of the nine types and on the official overall score.

\begin{table*}[t]
\centering\footnotesize
\caption{\textbf{VSTI-Bench test split, accuracy per type (\%).} CS: \CS; SS: \SSt\ as published; A: metric expert; B: calibrated \SSt; BTR: \ACR; $\pi_t$: router weight. On the object-relation types A and B were not run separately. Best per row in bold.}
\label{tab:vsti}
\setlength{\tabcolsep}{4pt}
\adjustbox{max width=\textwidth}{%
\begin{tabular}{@{}lcccccc@{}}
\toprule
Type & CS & SS & A & B & BTR & $\pi_t$\\
\midrule
\multicolumn{7}{@{}l}{\textit{Camera types (3,253 questions, 250 videos)}}\\
camera displacement & 21.25 & \textbf{25.38} & 22.79 & 24.10 & 24.60 & 1.000\\
camera movement direction & 42.35 & 42.49 & 42.35 & 42.49 & \textbf{43.17} & 0.984\\
camera--object absolute distance & 27.00 & 38.15 & 37.79 & 42.46 & \textbf{43.36} & 1.000\\
camera--object relative distance v1 & 69.35 & \textbf{79.03} & 69.35 & \textbf{79.03} & 69.35 & 0.000\\
camera--object relative distance v2 & 79.53 & 71.50 & 79.53 & 71.50 & \textbf{79.79} & 0.000\\
camera--object relative distance v3 & \textbf{83.41} & 74.53 & \textbf{83.41} & 74.53 & 83.11 & 0.000\\
Macro average, camera types & 53.82 & 55.18 & 55.87 & 55.69 & \textbf{57.23} & \\
\midrule
\multicolumn{7}{@{}l}{\textit{Object relations (1,314 questions, 186 videos)}}\\
object relative position, left/right & 64.27 & 70.23 & -- & -- & \textbf{71.05} & 1.000\\
object relative position, near/far & 53.35 & 61.83 & -- & -- & \textbf{62.50} & 1.000\\
object relative position, up/down & 81.27 & 87.34 & -- & -- & \textbf{87.60} & 1.000\\
Macro average, object relations & 66.30 & 73.13 & -- & -- & \textbf{73.72} & \\
\midrule
Official overall score (nine types) & 57.98 & 61.16 & -- & -- & \textbf{62.73} & \\
\bottomrule
\end{tabular}}
\end{table*}

\textbf{CV-Bench.} The test split has 1,278 single-image questions, and the stronger expert is \SSt. \ACR\ scores 82.99\%, which is 3.16\% above \SSt\ and 5.86\% above \CS.

\begin{table*}[t]
\centering\footnotesize
\caption{\textbf{\ACR\ against the baseline and the stronger expert on the test splits (\%).} CS: \CS; Stronger: the better expert on that split. VSTI-Bench is the macro average over six camera types. Differences are absolute, with paired 95\% intervals.}
\label{tab:summary}
\setlength{\tabcolsep}{3.5pt}
\adjustbox{max width=\textwidth}{%
\begin{tabular}{@{}lccccccc@{}}
\toprule
Benchmark, test & CS & Stronger & \ACR & Difference vs CS & CI & Difference vs stronger & CI\\
\midrule
VSI-Bench & 57.41 & 61.74 & 64.28 & +6.87 & [4.88, 9.15] & +2.54 & [0.66, 4.68]\\
VSTI-Bench, six camera types & 53.82 & 55.87 & 57.23 & +3.42 & [2.25, 4.63] & +1.36 & [0.10, 2.55]\\
CV-Bench & 77.13 & 79.83 & 82.99 & +5.86 & [3.50, 8.22] & +3.16 & [1.69, 4.69]\\
\bottomrule
\end{tabular}}
\end{table*}

\subsection{Is the gain due to non-visual shortcuts?}
VSI-Bench-Debiased~\citep{brown2026benchmark} removes the questions that a text-only model can answer from non-visual patterns. We re-score the existing predictions on its test part (1,160 questions), with no new run and no choice made on these questions. \CS\ scores 49.72\% and \ACR\ 56.46\%, so \ACR\ is 6.74\% above \CS\ and 4.69\% above \SSt. The gain over \CS\ is of similar size to that on the full test split (6.87\%), so it does not shrink when the non-visual shortcuts are removed. Table~\ref{tab:debiased} gives the scores per type: \ACR\ is the best or tied best on five of the eight types, with the largest gains over \CS\ on room size (+26.57\%) and relative direction (+10.47\%).

\begin{table}[t]
\centering\footnotesize
\caption{\textbf{VSI-Bench-Debiased, test part, accuracy per type (\%).} 1,160 questions; existing predictions re-scored. CS: \CS\ official; A: metric expert; SS: \SSt\ as published; BTR: \ACR. Best per row in bold.}
\label{tab:debiased}
\setlength{\tabcolsep}{4pt}
\begin{adjustbox}{max width=\columnwidth}
\begin{tabular}{@{}lcccc@{}}
\toprule
Type & CS & A & SS & BTR\\
\midrule
Appearance order & 76.65 & 79.04 & 76.05 & \textbf{85.03}\\
Absolute distance & 38.21 & 42.80 & 42.29 & \textbf{44.54}\\
Object counting & 50.88 & \textbf{51.68} & 48.00 & \textbf{51.68}\\
Relative direction & 48.35 & 51.69 & \textbf{59.67} & 58.82\\
Relative distance & 59.54 & \textbf{61.07} & 57.25 & 60.31\\
Object size & 52.10 & \textbf{54.52} & 41.78 & \textbf{54.52}\\
Room size & 45.25 & 67.68 & 64.14 & \textbf{71.82}\\
Route planning & \textbf{26.79} & 25.00 & 25.00 & 25.00\\
\midrule
Macro average & 49.72 & 54.19 & 51.77 & \textbf{56.46}\\
\bottomrule
\end{tabular}
\end{adjustbox}
\end{table}

\section{Limitations and future work}
\ACR\ runs both experts on every question, and its router needs question-type labels and a small labelled development set. Skipping the second expert when $\pi_t$ is 0 or 1, predicting the question type with a classifier, and routing per question instead of per type are natural next steps.

\section{Conclusion}
\ACR\ assigns spatial tasks to two frozen VLMs according to their abilities. A metric expert, \CS\ with training-aligned sampling, metric-aware minimum Bayes risk decoding and per-type metric calibration, estimates sizes and distances; a geometry expert, \SSt, handles relations and directions; and a Bayesian task router decides, for each question type, how much to rely on each. Without training any model weight, \ACR\ raises \CS\ from 57.41\% to 64.28\% on the VSI-Bench test split, beats the stronger expert on VSI-Bench, the camera questions of VSTI-Bench and CV-Bench, and outperforms the best published small VLMs by 3.17\% on VSI-Bench and, excluding a model trained on CV-Bench-style data, by 1.44\% on CV-Bench.

\bibliography{refs}

\end{document}